\documentclass[sigconf,screen]{acmart}

\usepackage{xcolor}
\usepackage{graphicx}
\usepackage{subcaption}
\usepackage{bm}
\usepackage{bbm}
\usepackage{dsfont}
\usepackage{xspace}
\usepackage{url}
\usepackage{multirow}
\usepackage{booktabs}
\usepackage{enumerate}
\usepackage[inline]{enumitem}
\usepackage{acronym}
\usepackage{balance}
\usepackage{mleftright}
\usepackage{float}
\usepackage{tikz}
\usetikzlibrary{positioning,arrows.meta}
\usepackage{enumerate}

\AtBeginDocument{%
  }

\newcommand{\ie}{\emph{i.e.,}\xspace}

\newcommand{\eg}{\emph{e.g.,}\xspace}

\newcommand{\ignore}[1]{}

\newcommand{\hlcauto}[2]{%
  \begingroup
  \setlength{\fboxsep}{1pt}%
  \raisebox{0ex}{\colorbox[HTML]{#1}{#2}}%
  \endgroup
}

\definecolor{PreferredGreen}{RGB}{34,139,34}

\acrodef{IR}{information retrieval}
\acrodef{RS}{recommender system}
\acrodef{NLP}{natural language processing}
\acrodef{ML}{machine learning}
\acrodef{CV}{computer vision}
\acrodef{LLM}{large language model}
\acrodef{VLM}{vision language model}
\acrodef{VLAM}{vision-language-action model}
\acrodef{CF}{collaborative filtering}
\acrodef{CB}{content-based}

\acrodef{HRI}{human robot interaction}
\acrodef{HCI}{human computer interaction}

\acrodef{DCG}{Discounted Cumulative Gain}

\acrodef{RL}{reinforcement learning}
\acrodef{RL4Rec}{reinforcement learning for recommendation}
\acrodef{ID}{in-distribution}
\acrodef{OOD}{out-of-distribution}
\acrodef{Seq-Rec}{sequential recommendation}
\acrodef{EIB}{error-imputation-{\allowbreak}based model}
\acrodef{IPS}{inverse propensity scoring}
\acrodef{PWE}{propensity-weighted estimator}
\acrodef{DR}{doubly robust}
\acrodef{SNIPS}{self-normalized inverse propensity scoring}
\acrodef{MNAR}{missing not at random}
\acrodef{Pop}{popularity}
\acrodef{Avg}{average}
\acrodef{MF}{matrix factorization}
\acrodef{BPR}{Bayesian personalized ranking}
\acrodef{TMF}{time-aware matrix factorization}
\acrodef{TTF}{time-aware tensor factorization}
\acrodef{TMTF}{time-aware matrix \& tensor factorization}
\acrodef{MC}{Markov chains}
\acrodef{DQN}{Deep Q-Network}
\acrodef{DQN4Rec}{Deep Q-Network based recommendation}
\acrodef{MDP}{Markov decision process}
\acrodef{GT}{Ground Truth}
\acrodef{sim-GT}{simulated Ground Truth}
\acrodef{MCAR}{missing completely at random}
\acrodef{RNN}{recurrent neural network}
\acrodef{GRU}{gated recurrent unit}
\acrodef{CNN}{convolutional neural network}
\acrodef{MLP}{multi-layer perceptron}
\acrodef{LSTM}{long short-term memory}
\acrodef{BOI}{bag of items}
\acrodef{PLD}{pairwise local dependency between items}

\acrodef{NLL}{Negative Log-Likelihood}
\acrodef{PPL}{Perplexity}

\acrodef{P}{Precision}
\acrodef{R}{Recall}
\acrodef{MAP}{Mean Average Precision}
\acrodef{MRR}{Mean Reciprocal Rank}
\acrodef{NDCG}{normalized discounted cumulative gain}

\acrodef{MSE}{mean squared error}
\acrodef{MAE}{mean absolute error}
\acrodef{RMSE}{root mean square error} 
\acrodef{ALS}{alternating least squares}

\newcommand{\header}[1]{\vspace*{1mm}\noindent\textbf{#1}.}

\setcopyright{cc}
\setcctype{by}
\acmDOI{10.1145/3757279.3785610}
\acmYear{2026}
\copyrightyear{2026}
\acmISBN{979-8-4007-2128-1/2026/03}
\acmConference[HRI '26]{Proceedings of the 21st ACM/IEEE International Conference on Human-Robot Interaction}{March 16--19, 2026}{Edinburgh, Scotland, UK}
\acmBooktitle{Proceedings of the 21st ACM/IEEE International Conference on Human-Robot Interaction (HRI '26), March 16--19, 2026, Edinburgh, Scotland, UK}
\received{2025-09-30}
\received[accepted]{2025-12-01}



\author{Jin Huang}
\orcid{0000-0001-9273-9037}
\affiliation{%
  \institution{University of Cambridge}
  \city{Cambridge}
  \country{United Kingdom}
}
\email{jh2642@cam.ac.uk}

\author{Fethiye Irmak Doğan}
\orcid{0000-0002-1733-7019}
\affiliation{%
  \institution{University of Cambridge}
  \city{Cambridge}
  \country{United Kingdom}
}
\email{fid21@cam.ac.uk}

\author{Hatice Gunes}
\orcid{0000-0003-2407-3012}
\affiliation{%
  \institution{University of Cambridge}
  \city{Cambridge}
  \country{United Kingdom}
}
\email{hg410@cam.ac.uk}

\acmSubmissionID{3751}

\begin{document}

\title{Reimagining Social Robots as Recommender Systems: Foundations, Framework, and Applications}

\titlenote{%
This work was supported by the EU’s Horizon Europe research and innovation programme under the Marie Skłodowska-Curie Actions Postdoctoral Fellowships (European Fellowship) 2024, grant agreement no. 101203728 — SOCIALADAPT — HORIZON-MSCA-2024-PF-01. The works of F. I. Doğan \& H. Gunes were supported in part by CHANSE \& NORFACE through the MICRO project, funded by ESRC/UKRI (grant ref. UKRI572).
Views and opinions expressed are those of the author(s) only and do not necessarily reflect those of the funding bodies. Neither the European Union nor the granting authorities can be held responsible for them.
\textbf{Contributions:} Conceptualisation \& Funding acquisition: HG, JH. Methodology \& writing: JH, FID, HG. Supervision \& project administration: HG. 
}

\begin{abstract}
  Personalization in social robots refers to the ability of the robot to meet the needs and/or preferences of an individual user. 
  Existing approaches typically rely on \acp{LLM} to generate context-aware responses based on user metadata and historical interactions or on adaptive methods such as \ac{RL} to learn from users' immediate reactions in real time.
  However, these approaches fall short of comprehensively capturing user preferences--including long-term, short-term, and fine-grained aspects--, and of using them to rank and select actions, proactively personalize interactions, and ensure ethically responsible adaptations.
  To address the limitations, we propose drawing on \acp{RS}, which specialize in modeling user preferences and providing personalized recommendations. 
  To ensure the integration of \ac{RS} techniques is well-grounded and seamless throughout the social robot pipeline, we (i)~align the paradigms underlying social robots and \acp{RS}, (ii)~identify key techniques that can enhance personalization in social robots, and (iii)~
  design them as modular, plug-and-play components.
  This work not only establishes a framework for integrating RS techniques into social robots but also opens a pathway for deep collaboration between the RS and HRI communities, accelerating innovation in both fields.
\end{abstract}

\maketitle

\acresetall

\section{Introduction}

Social robots are commonly defined as robots that can execute designated tasks while interacting with humans by adhering to certain social cues and rules~\cite{yan2014survey}. 
Their ability to interact socially enables them to provide companionship, emotional support, and assistance with daily tasks to various user groups, %
thereby playing a crucial role in responding to users' socioemotional needs~\cite{herse2018you}.
Research on social robots is inherently complex and interdisciplinary, drawing on fields such as computer vision, psychology, \acl{NLP}, and mechatronics.
A typical processing pipeline of social robots includes: a \emph{robot perception} module that acquires environment information and translates it into semantic understanding, a \emph {robot cognition} module that connects perception to action decision-making, and a \emph{robot action} module that executes behaviors through various modalities~\cite{yan2014survey}.
The rapid progress of foundation models, especially the widespread adoption of \acp{LLM}, has endowed social robots with strong capacities for language understanding, instruction following, and dialogue generation in the robot cognition module~\cite{obrenovic2024generative,zhang2023large}. Moreover, \aclp{VLAM} allow robots to directly perceive the environment, understand complex instructions, and execute appropriate actions dynamically~\cite{zitkovich2023rt}, thereby reducing reliance on a distinct robot cognition component.

A key factor in enhancing user satisfaction and experience with social robots is \emph{personalization}, which broadly refers to the ability of a robot to meet the needs and/or preferences of an individual user~\cite{gasteiger2023factors}.
LLM-enhanced social robots can generate context-aware responses and adapt communication style to user characteristics by leveraging prompts that include user metadata (\eg demographics or role-based attributes) and historical interactions~\cite{huang2024selective,irfan2024recommendations}. 
Moreover, recent work employs \ac{RL} for real-time adaptation to users' immediate reactions, \eg gestures, facial expressions, or speech~\cite{holk2024polite,wang2024personalization}.
However, these approaches fall short of comprehensively capturing user preferences--which should include long-term, short-term, and fine-grained aspects--, and of using them to rank and select actions, proactively personalize interactions, and ensure ethically responsible adaptations.

\begin{figure*}[t]
    \centering
    \includegraphics[width=0.87\linewidth,alt={The figure illustrates the integration of Recommender System (RS) modules into the typical social robot flow.}, trim=10pt 10pt 15pt 15pt]{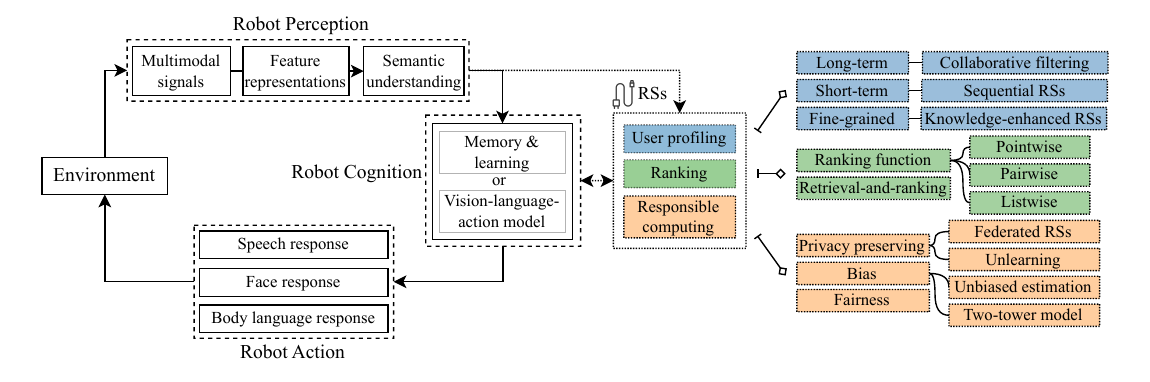} \vspace{-12pt}
    \caption{\small The typical HRI flow with plug-and-play RS modules. The RS modules enhance robot cognition through comprehensive user profile modeling, optimal action ranking, and responsible computing, mapping perceptual signals into personalized response choices.} \vspace{-13pt}
    \label{fig:general_flow}
    \Description{The figure illustrates the integration of Recommender System (RS) modules into the typical social robot flow.}
\end{figure*}

To address these limitations, it is natural to draw on \acp{RS}, which specialize in modeling user preferences and providing personalized recommendations, enabling RSs to filter relevant items from a large pool of candidates for individual users, thus mitigating information overload~\cite{bobadilla2013recommender}.
One predominant technical perspective in \acp{RS} is to formulate the recommendation task as a ``matrix filling'' problem, where the input to the \acl{ML} algorithm is a sparse user-item rating matrix and the output is the prediction of the missing ratings, an approach commonly referred to as \acf{CF}~\cite{jannach2021recommender,su2009survey}.
Building on this, deep learning algorithms have been extensively adopted in \acp{RS} to improve recommendation accuracy~\cite{zhang2019deep}, introduce new RS task formulations--\eg using \ac{RNN} to model short-term preferences given sequential user interaction sessions~\cite{quadrana2018sequence,wang2021survey}--, and drive advancements, including memory networks integrated with structured knowledge to capture fine-grained user preferences and improve explainability~\cite{chen2018sequential,huang2018improving}.
Moreover, a branch of research extends beyond accuracy, addressing bias and promoting fairness and diversity to mitigate negative effects of RSs, \eg filter bubbles~\cite{chen2023bias}; while other efforts explore federated learning~\cite{asad2023comprehensive} and unlearning approaches~\cite{li2024survey} to safeguard user privacy.

We argue that \ac{HRI} and social robots can benefit significantly from \acp{RS}, \eg for improving personalization and user experience.
However, limited research has explored this intersection. 
The only relevant study~\cite{swearingen2001beyond} examined user interactions with online book and movie \acp{RS} from a human-computer interaction (HCI) perspective, but did not delve into how to implement \acp{RS} in \ac{HRI} or social robotics.
Surprisingly, topics featured in the ACM conference Series on \acl{RS}~\cite{recsyshome} and the ACM/IEEE International Conference on Human-Robot Interaction~\cite{hri2025cfp} reveal no commonality. 
Still, the ACM Transactions on Recommender Systems~\cite{torsguidelines} explicitly encourages submissions that investigate HCI aspects, providing an initial step in this direction.
To take this further, this paper investigates what social robots as \acp{RS} would look like, why this matters, and how it could be implemented.

We first revisit and align the design paradigms underlying \acp{RS} and social robots.
We then identify key techniques of RSs that can enhance the personalization of social robots and investigate how they can be appropriately and flexibly integrated into the social robot pipeline.
The key techniques include: 
(i) \emph{user profiling techniques} for comprehensively modeling long-term, short-term, and fine-grained user preferences;
(ii) \emph{ranking techniques} for efficiently determining optimal item ordering; and
(iii) \emph{responsible computing techniques} for safeguarding user privacy, mitigating bias, and promoting fairness.
These techniques are implemented as modular, plug-and-play components that integrate seamlessly throughout the social robot pipeline (See Figure~\ref{fig:general_flow}).
We provide mathematical formulations and use cases to demonstrate the feasibility and efficacy of the proposed framework, identify key open challenges, and suggest future research directions.
Our proposed framework not only integrates RSs with social robots, but also bridges the RS and HRI communities, accelerating innovation and enabling collaborative research initiatives across both disciplines.

\section{Conceptualization}
\label{sec:concept}
This section revisits and aligns the paradigms underlying \acp{RS} and social robotics to build a conceptual foundation for their integration.

\header{The Social Robotic Paradigms}
A social robot is generally defined as an autonomous robot that can execute designated tasks and interact with humans by adhering to certain social cues and rules~\cite{yan2014survey,bartneck2004design,sarrica2020many}.
The design of social robots is an interdisciplinary undertaking, divided into different paradigms~\cite{frieske2024survey}:
\vspace{-2pt}
\begin{enumerate}[leftmargin=13pt]
    \item \emph{Cognitive architectures} (\eg that follow the standard model of the mind) are based on insights from neuroscience, providing a framework designed to capture general cognitive functions, including perception, memory, learning, and decision-making. 
    \item \emph{Role design model} establishes predefined roles and actions for both robots and humans to complete tasks, rather than accommodating flexible user actions.
    \item \emph{Linguistic model} accommodates the initiative for communication from both users and robots, categorizing speech actions into giving and receiving information.
    \item \emph{Communication flow}
    provides a framework for managing the exchange of information in a dialog to ensure interactions are coherent, contextually relevant, and goal-oriented.
    \item \emph{Activity system model} considers the broad context of activities and social interactions, emphasizing the role of mediating tools (\eg language, gestures, artifacts) between humans and robots, facilitating communication and task completion.
    \item \emph{Integrate design model} focuses on a holistic design of multimodal dialogue systems for social robots by combining the operational (task performance), emotional (emotion recognition/response), and communicational (dialogue management) dimensions.
    \item \emph{Evaluation paradigm} combines quantitative methods (task success rates, error rates) and qualitative methods (interviews, surveys, observations) to assess complex robotic dialogue systems.
\end{enumerate}

\begin{table*}[ht!]
    \centering
    \caption{\small Alignment of paradigms between social robotics and \acp{RS} for advanced personalized social robots. \emph{Bridges} column shows which RS techniques enable each functional alignment: UP for user profiling, R for ranking, RC for responsible computing, and ALL involving all three.} \vspace{-12pt}%
    \small
    \begin{tabular}{| p{2cm} | p{2.6cm} | p{2.6cm} | p{8cm} | p{0.8cm} |}
        \toprule
        Theme & Social robotics & \acp{RS} & Functional alignment & Bridges\\
        \midrule
        {User modeling \newline (Robot Cognition)} & Cognitive architectures, role design model & Representation paradigm, methodological paradigm & User profiles and representations dynamically gathered and created using RS techniques, augmented by broad data and knowledge, enable robots to understand users' preferences and needs dynamically. & UP\\ 
        \midrule
        Interaction \newline (Robot Cognition \& Robot Action) & Linguistic model, communication flow & Interaction paradigm & Interaction strategies in social robots are enhanced by foundation model generation and personalized/adapted through RS-based profiling and ranking, while mitigating bias and ensuring safety. & ALL \\
        \midrule
        System design \newline (Whole flow) & Activity system, integrate design model & Methodological paradigm & Inferred user preferences can guide the robot response ranking and selection using RS techniques. & UP \& R \\
        \midrule
        Evaluation \newline (Whole flow) & Evaluation paradigm & Performance assessment, ethical considerations & Evaluation combines qualitative and quantitative approaches with personalization quality assessment, incorporating ranking-aware RS metrics and LLM annotations, while considering responsibility aspects. & R \& RC \\ 
        \bottomrule
    \end{tabular} \vspace{-12pt}
    \label{tab:RS4SR:paradigms_alignment}
    \Description{Alignment of paradigms between social robotics and \acp{RS} for advanced personalized social robots.}
\end{table*}

\noindent\textbf{The Recommender System Paradigms.}
Unlike social robotics, which often involves complex physical interactions and multimodal interpretation of social cues, \acp{RS} typically address a more well-defined task, which predicts user preferences and generates personalized item suggestions, making them a key area of study in the application of machine learning~\cite{bobadilla2013recommender,zhang2019deep}.
The field can be understood through the following key paradigms:
\begin{enumerate}[leftmargin=13pt]
    \item \emph{Representation paradigm} defines how users and items are internally represented in RSs. The most widely adopted is the ID-based paradigm, in which users and items are represented as discrete identifiers; this has dominated \acused{RS}\acp{RS} for over a decade~\cite{li2023exploring}. Alternatively, users and items can be represented via descriptive features (\eg meta-data, attributes, temporal signal), multiple modalities (\eg textual descriptions, interaction histories, contextual signals, external knowledge base), or hybrid approaches.
    \item \emph{Methodological paradigm} provides frameworks for predicting user preferences and making recommendations. 
    User preference modeling can be categorized as \acf{CF}~\cite{su2009survey}, which relies on interaction patterns among similar users or items; \acl{CB}~\cite{pazzani2007content}, which recommends items similar to those the user liked in the past; and hybrid approaches that combine both. 
    Additionally, various ranking functions have been applied to optimize recommendation quality and robustness.
    \item \emph{Interaction paradigm} concerns how \acp{RS} engage with users. 
    Most \acp{RS} assume single-turn settings, while RSs can adopt multi-turn~\cite{zhao2024let,huang2020keeping} or conversational approaches~\cite{jannach2021survey}, which iteratively adapt recommendations based on immediate feedback.
    \item \emph{Performance assessment} defines the optimization objectives and evaluation methods for RSs. Traditional approaches prioritize accuracy (\eg precision and recall of relevant item predictions), typically measured via offline evaluation using logged data, though this suffers from distribution shift issues~\cite{canamares2020offline}.
    An alternative is online A/B testing in the platform with actual users, which provides a reliable evaluation but is costly and risky~\cite{beel2015comparison}.
    \item \emph{Ethical considerations.} Core concerns include bias and fairness, to prevent systematic disadvantages for particular user and/or item groups~\cite{chen2023bias}; privacy, to protect sensitive personal data from misuse or leakage~\cite{jeckmans2012privacy}; and transparency and explainability, to make recommendation processes more interpretable to users and accountable to stakeholders~\cite{wang2024trustworthy}.
    Neglecting these can lead to negative effects such as filter bubbles, unfairness, privacy risks, and undermining trust due to opaque decision-making, particularly problematic in high-risk domains, \eg news \acp{RS}.
\end{enumerate}

\noindent\textbf{Alignment between \ac{RS} and social robotics paradigms}.
Social robotics paradigms have been translated into a robot's practical functioning.
For instance, the cognitive architectures paradigm translates into the perception-cognition-action loop~\cite{yan2014survey}, the typical pipeline that governs robot interaction with its environment and with users.
Within this pipeline, we systematically align paradigms from both domains for the functional integration of RSs with social robots -- see Table~\ref{tab:RS4SR:paradigms_alignment}.
The alignment spans four key themes: user modeling, interaction, system design, and evaluation. For each theme, we map corresponding paradigms in social robotics and RSs, describe their functional alignment, and specify which RS techniques (\ie user profiling, ranking, and responsible computing, further detailed in Section~\ref{sec:socialrobots_as_rs}) serve as bridges to enable this integration. This systematic alignment provides a conceptual foundation for transferring RS methodologies to enhance social robot personalization while maintaining ethical and safety considerations.

\header{Differences between social robotic and \ac{RS} paradigms}
Despite the shared goal of personalization, social robotics and \acp{RS} differ in how they operationalize it.
Current social robots, most enhanced by using foundation models (\eg \acp{LLM}), can achieve personalization through two prominent strategies: incorporating user metadata and interaction history in \ac{LLM} prompts to generate personalized responses, and applying reinforcement learning to reactively adapt robot behavior based on real-time user feedback~\cite{holk2024polite,wang2024personalization}.
In contrast, RSs systematically model comprehensive user preferences at contextual and individual levels, efficiently determine optimal item ranking, and safeguard user privacy while mitigating bias and promoting fairness.
These capabilities (user profiling, ranking, and responsible computing) are the core techniques we bring from RSs to social robots, with their functional integration detailed in Section~\ref{sec:socialrobots_as_rs} and use cases presented in Section~\ref{sec:use-cases}.

\section{Recommender Systems for Social Robots}
\label{sec:socialrobots_as_rs}
This section details how \ac{RS} techniques are functionally integrated into social robots.
Our integration follows three principles:
(i) Modularity: RS components should be plug-and-play, enabling easy swapping and testing without re-engineering the entire system~\cite{mcdc2025}.
(ii)~Human-in-the-loop: Integration should allow experts, stakeholders, and users to provide corrections throughout the system lifecycle~\cite{bisen2020human}.
(iii) Transparency: RS integration should be interpretable and explainable to users~\cite{edwards2021eu}.
Accordingly, RS techniques are implemented as modular components with well-defined interfaces,
ensuring reproducibility and extensibility.

\subsection{General framework for Integration}
\noindent\textbf{Processing pipeline of social robots.}
Social robots are typically deployed to engage with human users, processing multimodal environment signals $e \in E$ to deliver contextually appropriate robot actions $a \in A$.
The interaction pipeline includes three key modules: 
(i)~Perception module $\mathcal{P}: E \rightarrow O$, which captures and processes environmental and human signals into features and semantic representations $O$ through multimodal sensor fusion and deep learning models.
(ii)~Cognition module $\mathcal{C}: O \times S \rightarrow S' \times D$, which integrates the current observations from the environment with the robot's memory $S$ to generate an updated memory $S'$ and a decision representation $D$ that encodes potential action options.
(iii)~Robot action module $\mathcal{A}: D \rightarrow A$, which, based on decision representation $D$, selects and executes the most appropriate actions such as tailored responses or physical behaviors, ensuring seamless and user-centric human-robot interaction.
Social robots interact with human users through multiple iterative cycles, a process referred to as a feedback loop, where historical interactions are updated and stored in the robot's memory state~$S'$.
Building on recent foundation model advances, instead of relying on separated modules, \acfp{VLAM} can directly map environmental observations to executable robot actions, $\text{VLAM}: E \times S \rightarrow S' \times A$.

\header{Integration of RS with social robots}
RS methods offer several key advantages: comprehensive user preference modeling, efficient item ordering, and privacy safeguards with bias mitigation and fairness promotion. 
To leverage these advantages, we identify and integrate three key RS techniques into social robots:
\begin{enumerate}[leftmargin=13pt]
    \item \emph{User profiling techniques:} $\mathcal{U}: O \times U \rightarrow U'$, where environmental observations $O$ will be used to update the current user profile~$U$ into an updated profile $U'$.
    \item \emph{Ranking techniques} covering: (i) Ranking functions $\mathcal{R}: U \times A \rightarrow R$, where user profile $U$ is used to compute personalized relevant scores $R$ on robot actions $A$ given the corresponding user; and 
    (ii) retrieval-and-reranking framework: first retrieve relevant candidate actions $\mathcal{R}_\text{retrieval}: O \times U \rightarrow A_\text{candidates}$ and then rerank these candidates $\mathcal{R}_\text{rerank}: U \times A_\text{candidates} \rightarrow R_\text{candidates}$.
    $R_\text{candidates}$ will enhance $D$ and guide action selection.
    \item \emph{Responsible computing techniques} for safeguarding user privacy, mitigating bias, and promoting fairness. 
    One simple way to formulate this is as constraints on existing operations, \eg fairness-aware ranking: $\mathcal{R}_\text{fairness}: U \times A \rightarrow R \text{ subject to}\,\, \mathcal{C}_\text{fairness}$, ensuring ethical behavior during normal system operation.
    Other approaches necessitate separate operational modes, \eg profile unlearning: $\mathcal{U}_\text{unlearn}: U \times O_\text{forget} \rightarrow U'$, fundamentally altering system behavior for selective data $O_\text{forget}$ removal rather than following existing operations.
\end{enumerate}
\vspace{-2pt}
Figure~\ref{fig:general_flow} summarizes three key RS techniques for social robots and their taxonomy, with implementation details in later subsections.

\subsection{User Profiling Techniques}
\label{sec:profiling}
This section reviews user profiling techniques from \acp{RS} for social robots across different granularities.

\header{Granularity at user level}
Personalization in social robots can be categorized into three levels: global-level, group-level (also referred to as persona-level), and individual-level. 
Global-level reflects general preferences and norms that are widely accepted across a broad population.
Group-level targets groups of users who share similar characteristics or personas.
These two levels have been extensively studied in psychology and \acl{HRI}, providing the foundation for widely-adopted rule-based and expert-designed frameworks in social robots~\cite{hellou2021personalization,gasteiger2023factors,rossi2017user}, and mitigating cold-start issues in early interactions with new users~\cite{zhang2025cold}.

Explicit user profile data provided by users is commonly used to identify a user's group for group-level personalization~\cite{purificato2024user,gasteiger2023factors}, such as:
(i) demographics, \eg age, gender, cultural background, language preferences; 
(ii) declared preferences, \eg preferred tone, types of conversation and interaction;
(iii) social context: relationship to robot (patient, student, companion), household composition; 
(iv)~activities and purposes, \eg therapeutic targets, learning objectives, entertainment needs.

In contrast, user historical interactions with robots and their feedback are more likely to reveal individual-level preferences, and \acp{RS} are particularly effective at automatically inferring individual preferences from these interactions.
Moreover, individual-level personalization can be aggregated to obtain group-level and global-level personalization~\cite{ye2025disentangled,cao2018attentive}.
When individual interaction data are sparse, RSs support individual-level cold-start and early adaptation via content-based and hybrid approaches, transfer learning, LLM-based semantic reasoning, and interview-based approaches, with bandit- or RL-based exploration enabling progressive personalization as feedback accumulates~\cite{zhang2025cold,mcinerney2018explore}.
Hence, throughout the paper, we focus on using \acp{RS} to model individual user preferences.

\header{Contextual granularity} %
In \acp{RS}, user profiling techniques model user preferences across different temporal and dimensional granularities, encompassing long-term preferences, short-term preferences, and fine-grained preferences.
\vspace{-5pt}
\subsubsection{Long-term preferences} 
It refers to modeling stable user interests that persist over time. The predominant technique in \acp{RS} is \acf{CF}~\cite{su2009survey}, which leverages historical user-item interactions across all users in the system to infer individual long-term patterns, typically through similarity measures (\eg user- or item-based CF) or, most commonly, matrix factorization~\cite{koren2009matrix}.
When applied to social robots, \ac{CF} operates on the set of all observed historical interactions in the system, denoted as $\mathcal{H} = \{(u,a,f) \mid (u, a) \in U \times A \text{ and } f \text{ was observed in the past}\}$. Notably, $\mathcal{H}$ is accumulated over time from the set of observed interactions $O$.
\ac{CF} models aim to infer user long-term preferences~$\bm{p}^\text{CF}_u$ and to optimize predictive accuracy of feedback by minimizing the following loss: 
\begin{equation}
    \mathcal{L}^\text{CF} = \sum_{(u,a,f) \in \mathcal{H}} \delta \bigl(f - \bm{p}^\text{CF}_u {\bm{q}^\text{CF}_a}^\top \bigr), %
\end{equation}
where ${\bm{q}^\text{CF}_a}$ denotes trainable or pre-defined action embeddings, and the metric $\delta$ measures the difference between prediction and ground-truth, \eg \ac{MSE}. 
Both user and action embeddings can be further enhanced by integrating deep learning networks, \eg \acl{MLP}~\cite{he2017neural}.
For simplicity, we omit the regularization term in the loss function throughout the paper.

\subsubsection{Short-term preferences} 
It refers to dynamic user interests that evolve rapidly based on recent interactions and immediate context. 
In \acp{RS}, tasks such as sequential recommendations and session-based recommendations~\cite{quadrana2018sequence,wang2021survey} are specifically designed to capture these preferences.
In the case of a social robot, sequential \acp{RS} operate on temporally ordered interaction sequences for each user, denoted as $\mathcal{S}_u = \{(a_t, f_t) \mid t\in [1, T_u] \}$, where $T_u$ represents the length of user $u$'s interaction sequence. The interaction sequence $\mathcal{S}_u$ aggregates past observations $O$ in temporal order or labeled with timestamps.
Sequential \ac{RS} models infer user short-term preferences $\bm{p}^\text{Seq}_{u,t}$ at time $t$ by minimizing:
\vspace{-0.4em}
\begin{equation}
    \begin{split}
        \mathcal{L}^\text{Seq} &= \sum_{u} \sum_{t = 1}^{T_u} \delta \bigl(f_{t}, \bm{p}^\text{Seq}_{u,t} {\bm{q}^\text{Seq}_{a_t}}^\top \bigr), \\
        \bm{p}^\text{Seq}_{u,t} &= \text{Seq-}\mathcal{M}(\bm{p}^\text{Seq}_{u,t-1}, \bm{q}^\text{Seq}_{a_{t-1}}, f_{t-1}),\,\, t = 2, 3, \cdots, T_u
    \end{split} \label{eq:seq-rec}
\end{equation}
where $\text{Seq-}\mathcal{M}$ is typically instantiated with recurrent neural networks~\cite{tealab2018time} such as GRU~\cite{cho2014properties} and can be extended to use attention-based architectures such as Transformers~\cite{vaswani2017attention}. The initial state of user preference $\bm{p}^\text{Seq}_{u,0}$ can be set as a trainable vector or initialized from long-term user embeddings.

\subsubsection{Fine-grained preferences} 
It refers to user preferences across multiple dimensions or attributes of items, enabling the modeling of nuanced preferences beyond coarse item-level ratings. In \acp{RS}, effective approaches leverage structured knowledge associated with items, such as knowledge graphs for movies~\cite{wang2018ripplenet,huang2018improving} and taxonomies for products~\cite{huang2019taxonomy}, to capture these fine-grained aspects. 
In the context of social robots, building on sequential \acp{RS} as an example, knowledge-enhanced \acp{RS} operate on the temporal interaction sequences $\mathcal{S}_u$ and enrich each action $a$ with associated knowledge representations $\bm{k}_a$. These knowledge representations may include semantic categories (\eg greeting, entertainment, education), environmental contexts (\eg home and in the wild), interaction modalities (\eg verbal, gestural), social functions (\eg ice-breaking, emotional support), or domain-specific purposes (\eg therapeutic goal, cognitive load requirements).
Knowledge-enhanced models infer user fine-grained preferences $\bm{p}^\text{KE}_{u,t}$ by aligning sequential patterns with knowledge representations:
\begin{equation}
        \bm{q}^{\text{KE}}_{a_t} = \bm{q}^{\text{Seq}}_{a_t} \circ \bm{k}_{a_t}, \,\,\,\,
        \bm{p}^{\text{KE}}_{u,t} = \text{Seq-}\mathcal{M}(\bm{p}^{\text{KE}}_{u,t-1}, \bm{q}^\text{KE}_{a_{t-1}}, f_{t-1}).
\end{equation}
Replacing $\bm{q}^{\text{Seq}}_{a_t}$ and $\bm{p}^{\text{Seq}}_{u,t}$ with above knowledge-aligned representations, all trainable parameters can be optimized following Eq.~\ref{eq:seq-rec}.

\subsection{Personalization Ranking}
\subsubsection{Rank functions}
In \acp{RS}, users are typically presented with a ranked list of items and provide their feedback through their selections, which also reflects their preferences over the ranking order~\cite{swaminathan2017off,jalili2018evaluating}.
In contrast, social robots usually execute only the top-1 ranked action and receive user feedback solely on that action.
Despite this, robots can maintain the full ranking list or associated relevant scores $R_\text{candidates}$ for all possible candidate actions in their decision representation $D$. This enables users to provide feedback on the ranking during follow-up interactions, for instance, by prompting ``\emph{can you do cleaning first?}'', which signals positive feedback on the cleaning action even if it was not the top-1 action decision.
Thus, the use of ranking-aware functions is also beneficial for optimizing user preferences in social robots.

Formally, the ranking list is obtained by ordering actions according to their predicted relevance scores $\hat{r}_{u,a} = \bm{p}_u \bm{q}_a^\top$, where user preference $\bm{p}_u$ can be any type of preferences defined in Section~\ref{sec:profiling} or their combinations.
To leverage ranking information in the optimization and evaluation of user profile modeling, \ac{RS} methods apply pairwise and listwise approaches--originally developed in the \acl{IR} domain~\cite{li2014learning,cao2007learning}--rather than the pointwise approach currently used in Section~\ref{sec:profiling}.
\begin{itemize}[nosep,leftmargin=10pt]
    \item \emph{Pairwise approach} focuses on relative ordering between interaction pairs~\cite{rendle2009bpr}. For user $u$, action $a$ is more relevant than $a'$, denoted as $(u,a) \succ (u,a')$, if user feedback $f$ on action $a$ is more positive than feedback $f'$ on action $a'$. The learning of user preferences can be enhanced by optimizing the following pairwise loss defined over the relative preferences of actions:
    \begin{equation}
        \mathcal{L}_\text{pairwise} = \sum_{(u,a) \succ (u,a')} \sigma(\hat{r}_{u,a} - \hat{r}_{u,a'}),
    \end{equation}
    where pairwise loss function $\sigma(\cdot)$ is typically chosen as either the Bayesian Personalized Ranking (BPR) loss $\sigma(x) = \log(1 + e^{-x})$ or the hinge loss $\sigma(x) = \max(0, 1 - x)$.
    \item \emph{Listwise approach} focuses on the entire ranking list of actions. For user $u$, social robots produce a ranked action list $A_u = [a_1, a_2, \ldots, a_n]$ with the corresponding user feedback scores $F_u = [f_1, f_2, \ldots, f_n]$ from observations.
    We use $A_u^*$ to denote the optimal ranking of these same actions for user $u$, where $A_u^*(i)$ denotes the action that should be placed at position $i$ if we sort the actions in descending order according to their actual feedback scores $F_u$.
    One common approach is a probability-based approach that optimizes the following loss to update user profile modeling~\cite{xia2008listwise}:
    \begin{equation}
        \mathcal{L}_\text{listwise}^{\text{prob}} = -\sum_{u} \sum_{i=1}^{|A_u^*|} \log \frac{\exp(\hat{r}_{u,A_u^*(i)})}{\sum_{j=i}^{|A_u^*|} \exp(\hat{r}_{u,A_u^*(j)})}.
    \end{equation}
    Alternatives are based on rank-aware metrics, such as the commonly used \acl{DCG} metric: $\lambda_\text{DCG}(A_u, F_u) = \sum_{i=1}^{|A_u|} \frac{2^{f_i} - 1}{\log_2(i + 1)}$.
    The objective of optimizing user preference modeling to generate $A_u$, is then formulated as either (i)~maximizing $\lambda_\text{DCG}(A_u, F_u)$ via policy gradient optimization methods~\cite{oosterhuis2021computationally}, or (ii)~minimizing a differentiable approximation of the discrepancy between $\lambda_\text{DCG}(A_u, F_u)$ and $\lambda_\text{DCG}(A^*_u, F_u^*)$, where $F_u^*$ is obtained by sorting $F_u$ in descending order~\cite{taylor2008softrank}. %
\end{itemize}
While listwise methods are theoretically superior due to directly optimizing the full ranking in \acl{IR} and \acp{RS}, pairwise methods have become the practical choice due to their simplicity, robustness, and stability in optimization~\cite{li2014learning,zhang2023dive}.

\subsubsection{Retrieval-and-ranking framework}
First-stage retrieval in \acp{RS} relies on three core techniques that work synergistically to enable efficient candidate retrieval $A_\text{candidates}$ from large-scale item spaces~$A$~\cite{covington2016deep,li2014learning}: 
(i)~representation learning, transforming users and items into dense vector representations in a shared latent space, which would be similar to the user preference embedding and item embeddings in Section~\ref{sec:profiling}, but usually prefer shallow networks \eg matrix factorization; 
(ii)~indexing mechanisms organize these representations into efficient search structures that enable fast approximate nearest neighbor search; and
(iii)~similarity calculation, measuring the relevance between users and items through a similarity function, \eg dot-product.
These techniques can be directly applied to social robots.
In addition, rule-based approaches may also be useful for initial retrieval. For example, given environmental signals such as `at home' and `after lunch', the system could prioritize actions related to relaxation activities or cleaning activities while filtering out high-energy or meal-related actions.

The retrieved action candidates from the first stage are subsequently ranked according to individual user preferences in the second ranking stage with more sophisticated personalization methods, enabling the robot to select the most suitable action for execution.

\subsection{Responsible Computing}
\subsubsection{Safeguarding user privacy} 
A widely adopted framework for preserving privacy in \acp{RS} is \emph{federated learning}, where a global \ac{RS} model is trained by aggregating local parameter updates (\eg gradients) across devices without collecting raw user data~\cite{asad2023comprehensive,ammad2019federated}. 
This framework provides a level of privacy, as user data never leaves the local devices.
To further enhance privacy and mitigate risks of sensitive information leakage from gradient updates, various techniques have been explored, including differential privacy, homomorphic encryption, anonymization, and pseudonymization~\cite{asad2023comprehensive}.
Federated \acp{RS}, combined with such privacy-preserving techniques, can be directly applied to user profiling methods in social robots.

Another promising research direction is \emph{recommendation unlearning}, focusing on effectively forgetting specific data (\eg sensitive data) that was used during model training~\cite{li2024survey,chen2022recommendation}. 
This is challenging as machine learning models can memorize training data, allowing traces to persist even after the removal of specific records.
Current approaches are largely based on active unlearning~\cite{li2024survey}.
When applied to social robots, the unlearning objective works on user profiling method losses and is formulated as:
\begin{equation}
    \mathcal{L}_\text{unlearn}(\theta) = - \mathcal{L}(\theta; O_\text{forget}) + \beta\mathcal{L}(\theta; O_\text{retain}),
\end{equation}
where $O_\text{forget}$ denotes data to be unlearned and $O_\text{retain}$ is the remaining data to preserve; $\beta$ controls the trade-off between forgetting target data and performance on remaining data.
Here, $\mathcal{L}$ can be instantiated with any loss functions described in Section~\ref{sec:profiling}.

\subsubsection{Mitigating bias and promoting fairness}
Consistent with other AI methodologies, \acp{RS} face bias problems that lead to negative effects, \eg over-specialization, filter bubbles, and unfairness~\cite{nguyen2014exploring,chen2023bias}.
To mitigate these issues, significant research efforts have focused on debiased user profiling and fairness in recommendations as primary approaches.
To ensure the user profiling in social robots is unbiased or debiased, a widely used approach is based on \ac{IPS} for causal inference~\cite{imbens2015causal}, which derives an unbiased estimation for a wide range of performance estimators, such as \ac{MSE} in rating prediction models~\cite{huang2024going,schnabel2016recommendations}.
For example, by integrating with \ac{IPS}, the pointwise CF model is modified to minimize:
\begin{equation}
    \mathcal{L}^\text{CF-IPS} = \sum_{(u,a,f) \in \mathcal{H}} \frac{\delta \bigl(f - \bm{p}^\text{CF}_u {\bm{q}^\text{CF}_a}^\top \bigr)}{b_{u,a}},
\end{equation}
where propensity $b_{u,a}$ denotes the probability that action $a$ was shown to user $u$ in observed historical interactions and is commonly estimated by using naive Bayes with maximum likelihood.
Here, IPS weighs each interaction inversely to their propensity, and, thereby, corrects for the over-presentation resulting from bias in the data.
Another debiasing method applies a two-tower model~\cite{zhao2019recommending}, with one modeling bias estimation and the other modeling user/item representations. 
Its main drawback is the difficulty of disentangling bias from true preferences in the absence of bias labels.

Fairness in \ac{ML} has been extensively studied~\cite{mehrabi2021survey}. As an application area of \ac{ML}, \acp{RS} have adopted and extended existing fairness approaches, while tailoring them to domain-specific definitions of fairness~\cite{wang2023survey}. Fairness research in HRI~\cite{claure2022fairness} and social robots~\cite{cheong2024small} similarly adapts general \ac{ML} notions of fairness to the specific challenges of human–robot interaction. 
As \acp{RS} have engaged with fairness concerns at a greater methodological maturity, 
their insights—such as
fair representation learning~\cite{zhu2020measuring} and long-term fairness~\cite{liu2020balancing}—provide valuable opportunities for further advancing fairness in social robotics and ensuring that personalization operates in a socially responsible manner.

\section{Use Cases}
\label{sec:use-cases}

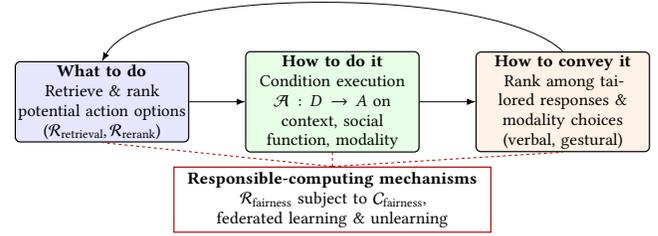
\begin{figure}[t!]
\centering
\vspace{-5pt}
\resizebox{\columnwidth}{!}{%
\begin{tikzpicture}[
every node/.style={font=\Huge},
  node distance=20mm,
  box/.style={rectangle, rounded corners=5pt, draw=black, very thick, text width=6cm, align=center, minimum height=1.2cm},
  arrow/.style={-{Latex[length=3mm]}, thick},
  constraint/.style={draw=red!70!black, very thick, text width=11cm, align=center, inner sep=5pt}
]

\node[box, fill=blue!10] (what) {\textbf{What to do}\\Retrieve \& rank potential action options\\($\mathcal{R}_{\text{retrieval}}, \mathcal{R}_{\text{rerank}}$)};
\node[box, fill=green!10, right=of what] (how) {\textbf{How to do it}\\Condition execution $\mathcal{A}: D \to A$ on context, social function, modality};
\node[box, fill=orange!10, right=of how] (convey) {\textbf{How to convey it}\\Rank among tailored responses \& modality choices (verbal, gestural)};

\draw[arrow] (what) -- (how);
\draw[arrow] (how) -- (convey);
\draw[arrow] (convey.north) to[out=135,in=60,looseness=0.5] (what.north);

\node[constraint, below=of how, yshift=15mm] (constraints) {\textbf{Responsible-computing mechanisms}\\
$\mathcal{R}_{\text{fairness}}$ subject to $\mathcal{C}_{\text{fairness}}$,\\ federated learning \& unlearning\\
};

\foreach \x in {what, how, convey} {
  \draw[dashed, red!70!black, thick] (constraints.north) -- (\x.south);
}

\end{tikzpicture}
}\vspace{-8pt}
\caption{\small RS-augmented HRI decision loop in action selection and response generation.} \vspace{-0pt}
\label{fig:decision-loop}
\Description{The figure illustrates RS-augmented HRI decision loop in action selection and response generation.}
\end{figure}

\acp{RS} in HRI can serve as a unifying mechanism that turns heterogeneous signals about a user, task, and environment into actionable choices for a robot’s perception, cognition, and action stack. 
Building on three pillars, (i) user profiling and state estimation (capturing long-term preferences, short-term intent, capabilities, and constraints), (ii) personalization and ranking (selecting and ordering candidate robot actions, content, or policies under uncertainty), and (iii) responsible computing (privacy, safety, fairness, and transparency by design),  RS modules can recommend 
\textbf{what to do} by retrieving and ranking the ``\emph{potential action options}'' encoded in $D$ ($\mathcal{R}_\text{retrieval}$, $\mathcal{R}_\text{rerank}$); determine \textbf{how to do it} by conditioning execution in $\mathcal{A}\!: D \!\rightarrow\! A$ on fine-grained attributes such as context, social function, and modality; and choose \textbf{how to convey it} by ranking among ``\emph{tailored responses}'' and modality choices (verbal, gestural). 
These decisions are constrained by responsible-computing mechanisms (\eg $\mathcal{R}_\text{fairness}$ subject to $\mathcal{C}_\text{fairness}$, federated learning, and unlearning) to balance utility with comfort, trust, and risk -- see Figure~\ref{fig:decision-loop}. 
Concretely, the proposed framework applies across multiple HRI domains, including socially aware service robots, collaborative robots, and companion robots; see Table~\ref{tab:scene:application} for illustrative examples.

\begin{table*}[t!]
    \centering
    \caption{\small Comparison of traditional, adaptive, and RS-enhanced approaches across common HRI applications. RS contributions align with three domains: \hlcauto{9BBEDB}{user profiling}, \hlcauto{A4D1A0}{personalized retrieval \& ranking}, and \hlcauto{FFCE9F}{responsible computing}, illustrating proactive and accountable personalization beyond static rules or short-horizon adaptation.} \vspace{-12pt}
    \small
    \begin{tabular}{| p{1.9cm} | p{2.6cm} | p{3.6cm} | p{8.2cm} |}
        \toprule
        Application & Traditional approaches & Adaptive approaches & What RSs bring \\
        \midrule

        Preference \& Priority Modeling 
        & One-size-fits-all policies; manual rules for typical users. 
        & Adaptation based on immediate feedback or a specific user~\cite{holk2024polite,wang2024personalization}. 
        & Uses \hlcauto{9BBEDB}{collaborative filtering} to model individuals and similar-user groups (e.g., safety-first, efficiency-first), enabling predictions of socially appropriate behaviors aligned with user needs. \\
        \midrule

        Contextual Action Selection 
        & Scripted actions tied to coarse context. 
        & Adjusts behavior from immediate implicit/explicit feedback or explanation~\cite{mcquillin2022learning,dougan2025grace, 10924711}. 
        & Learns \hlcauto{9BBEDB}{long-/short-term preferences} and group priorities (e.g., safety vs.\ efficiency); \hlcauto{A4D1A0}{ranks} candidate actions/variants to fit the user and scene. \\
        \midrule

        Turn-taking \& Interaction Pacing
        & Fixed turn order and timing; rigid initiative rules. 
        & Modulates timing based on immediate cues (e.g., response latency, gaze/attention)~\cite{ekstrom2022dual, 10731390}. 
        & Profiles \hlcauto{9BBEDB}{fine-grained} pacing and initiative preferences (per user/task); \hlcauto{A4D1A0}{ranks} turn-taking policies (e.g., wait time, interruption thresholds) to sustain engagement; applies \hlcauto{FFCE9F}{fairness}-aware calibration to avoid systematic over/under-interruption of specific user groups. \\
        \midrule

        Affective Expressiveness \& Feedback Style
        & Neutral or designer-defined affect; uniform feedback tone.
        & Modulates the robot's affective appraisal from immediate behavioural cues~\cite{10.3389/frobt.2022.717193}.
        & Learns \hlcauto{9BBEDB}{affect/verbosity sensitivities} and preferred feedback style; \hlcauto{A4D1A0}{ranks} expression modalities (speech, gesture, facial displays) to match user/context; enforces \hlcauto{FFCE9F}{fairness/transparency} for cross-cultural expressiveness.\\
        \midrule

        Collaboration Strategy \& Task Allocation
        & Static role assignment and task allocation. 
        & Allocates roles dynamically according to task or adjusts plans with user intentions~\cite{liu2024role, 9478205}.%
        & Uses \hlcauto{9BBEDB}{capability/pace profiles} and preferences to \hlcauto{A4D1A0}{rank} feasible strategies and allocations (assist level, role split) under safety/ergonomic constraints; supports mixed-initiative selection with \hlcauto{FFCE9F}{transparent}, auditable criteria. \\
        \midrule

        Goal/Entity Disambiguation 
        & Literal parsing; exhaustive search; frequent follow-ups. 
        & Requests clarification when ambiguity is detected~\cite{dougan2022asking, dogan2024model}. 
        & Applies \hlcauto{A4D1A0}{retrieval} of likely targets/locations from history and \hlcauto{A4D1A0}{reranking} conditioned on user-specific cues (e.g., ``my favorite mug''), pruning unfitting options and shrinking the search space. \\
        \midrule

        Conversational Intent \& Response Selection
        & Scripted replies and template matching.
        & Language models select from conversational history and content~\cite{leszczynski2025bt, kim2024understanding}.
        & Uses \hlcauto{A4D1A0}{retrieval-and-rerank} over candidate responses with \hlcauto{9BBEDB}{user/profile/history} and predicted intent; optimizes timing to align with preferences; supports \hlcauto{FFCE9F}{auditability} (explanations, safety filters).\\
        \midrule

        Proactive Assistance 
        & No routine awareness, or grounding from recognised action. 
        & Adapts based on contextual and environmental changes~\cite{patel2022proactive, bartoli2024streak}. 
        & Learns routine patterns with \hlcauto{9BBEDB}{sequential profiling} to \hlcauto{A4D1A0}{rank} proactive suggestions (e.g., likely next task), while enforcing \hlcauto{FFCE9F}{privacy safeguards} (federated learning, unlearning) when monitoring daily habits. \\
        \bottomrule
    \end{tabular} \vspace{-12pt}
    \label{tab:scene:application}
    \Description{This table illustrates the comparison of traditional, adaptive, and RS-enhanced approaches across common HRI applications.}
\end{table*}

\header{Socially Aware Service Robots}
Our framework can support socially aware service robots via several components. First, prior studies have explored the generation of socially appropriate robot actions (\eg vacuuming, carrying objects) in various contexts~\cite{churamani2024feature, 10.3389/frobt.2022.669420}. 
For instance, the \emph{robo-waiter}~\cite{mcquillin2022learning} enhanced sociability and appropriateness by incorporating real-time implicit (\eg facial affect) and explicit (\eg verbal) feedback. 
To tailor these actions to individual preferences, some approaches have leveraged personalized explanations~\cite{dougan2025grace}. 
In this context, user profiling techniques, such as collaborative filtering, can model individual preferences by leveraging patterns from user groups with shared priorities (\eg safety, efficiency), enabling accurate predictions of socially appropriate behaviors that align with users' needs. Similarly, robots designed to locate and retrieve objects in response to ambiguous commands~\cite{dougan2022asking, dogan2024model} can benefit from personalized ranking, particularly when queries include individualized preferences (\eg ``bring me my favorite mug''). 
While existing systems often rely on follow-up questions, retrieval and re-ranking techniques can eliminate unfitting locations and provide a more efficient alternative, reducing robots' search space. 
Finally, studies on proactive robot assistance~\cite{patel2022proactive, bartoli2024streak}, where robots observe user routines to anticipate needs (\eg if the user picks up a bowl, the robot offers cereal), would benefit from responsible computing practices, as safeguarding user data and privacy is essential when robots monitor daily habits and routines.

\header{Collaborative Robots}
Our framework can enhance collaborative robots in HRI through several complementary directions. First, studies on joint action and task allocation~\cite{petzoldt2023review, angleraud2021coordinating} have explored how robots share goals and coordinate with humans in activities such as assembly, carrying objects, or co-manipulating tools. 
User profiling techniques, such as clustering by work pace, physical ability, or communication style, can help robots anticipate preferred collaboration patterns and adapt their level of support accordingly. 
In another example, personalized ranking can refine how collaborative robots choose among multiple valid strategies by modeling a partner's preferred contribution to joint tasks~\cite{zhao2023learning}. 
Similarly, in a child–robot ``learning-by-teaching'' setting~\cite{ekstrom2022dual}, where a child tutors a questioning tutee-robot, RS-driven personalization adapts question difficulty, turn-taking, and calibrated displays of confusion/backchannels to sustain engagement and meet individual needs. 
Finally, responsible computing practices are essential when collaborative robots record motion data, monitor performance, or infer human intent~\cite{CAI2024102411}. 
Transparent data handling and privacy-preserving mechanisms foster user trust and ethical deployment.

\header{Companion Robots} 
Our framework can equip companion robots to model individual preferences, select and rank context-appropriate responses, and enforce responsible practices for emotionally sensitive exchanges.
First, systems that engage users in multimodal wellbeing conversations by integrating speech, gestures, and facial expressions to foster self-reflection and emotional support~\cite{10.1145/3712265} can draw on user profiling methods from \acp{RS} to tailor expressiveness, feedback style, and pacing to each individual’s preferences and sensitivities. 
Second, personalized ranking helps address challenges in LLM-powered conversational robots, \eg intent misinterpretation, untimely interruptions, or unacknowledged input~\cite{10.1145/3678957.3689030}. 
By evaluating candidate responses using predicted intent, context, and prior feedback, companion robots can select replies better aligned with user expectations. 
Finally, companion robots that support children's wellbeing~\cite{10.1145/2851581.2892409, borenstein2013companion} highlight the importance of responsible computing, \eg bias detection and fairness safeguards, to ensure transparent and equitable emotional guidance across diverse users.

\section{Open Challenges and Future Opportunities}
By incorporating \acp{RS}, social robots can solve some technical challenges that have long hindered effective human-robot interaction.
For instance, \citet{irfan2025between} identified persistent issues including:
(i) turn-taking problems such as frequent interruptions and slow responses,
(ii) repetitive responses,
(iii) superficial conversations, and
(iv) language barrier.
While they proposed rule-based solutions such as increasing silence timeout to account for long pauses in speech, these approaches require expert design for each specific case and lack generalization across different users and contexts.
In contrast, our proposed framework can automatically and adaptively solve these challenges by employing \acp{RS} techniques to learn individual user preferences and develop personalized interaction strategies that generalize across diverse users and contexts.

Building on this framework, we further identify key challenges and suggest potential directions for future research.

\header{Multi-user privacy control}
While our proposed framework integrates privacy-preserving RS techniques to safeguard user data, social robotics poses domain-specific privacy challenges~\cite{levinson2024our,bell2025always}. 
For example, household social robots serving multiple users (\eg children, older adults) face multi-user privacy challenges,
where individual preferences and data need protection from unauthorized access by other household members. Insufficient privacy controls could lead to unintended information sharing, violating privacy expectations and causing conflicts, particularly when vulnerable groups or sensitive information are involved. We suggest future work on hierarchical mask profiling systems with multi-layered privacy controls and privacy-preserving retrieval methods that filter robot actions based on user identity and access permissions.

\header{Integration and generalization}
Social robots increasingly rely on LLMs for natural language understanding and generation, often deployed via cloud-based infrastructures that reduce system-level integration complexity, introducing new challenges for integrating \ac{RS} modules with LLM-based cognition. A key problem is the mismatch between representation paradigms: \ac{RS} profiling modules generate dense user representations, while LLMs operate on text prompts. 
This poses bidirectional challenges for utilizing preferences within LLM-based cognition and incorporating LLMs into RS modules.
Social robots further require cross-domain preference generalization~\cite{zhu2022personalized}, where preferences learned in one context (\eg entertainment) generalize to other contexts (\eg education, therapy). 
We suggest future work on preference-grounded \acp{LLM} that incorporate learned preferences via retrieval-augmented generation or specialized prompting, \ac{LLM}-enhanced preference learning leveraging \ac{LLM} representations to capture user signals from conversations, and cross-domain preference transfer via shared representation and transfer learning.
Future work should also explore action representations for social robots, including continuous action spaces, that bridge learned preferences and executable behaviors.

\header{Contextual understanding}
Compared to LLMs relying on text and \acp{RS} relying mainly on interaction data, social robots operate in rich, multi-modal environments containing implicit user feedback including behavioral signals (\eg nonverbal cues, speech patterns), physiological responses (\eg heart rate, stress indicators), performance metrics (\eg task completion, learning progress), and emotional state (\eg mood, stress level). Current RS modules and foundation models struggle to grasp the intricate nuances of human-robot interactions~\cite{zhang2023large}. 
We suggest future work to include implicit feedback understanding that leverages multi-modal modeling to extract and fuse user signals from behavioral cues and context, and multi-modal preference learning that integrates these signals into \ac{RS} modules and LLMs to enhance user preference representations.

\header{Bias and fairness in embodied AI} 
Bias in \acp{RS} primarily focuses on item over-representation and user selection patterns. However, social robots face unique embodiment-related biases (\eg physical appearance, voice, movement, interaction styles)~\cite{liu2025roboviewbiasbenchmarkingvisualbias} and natural language-related biases from \ac{LLM} training corpora, manifesting through biased text generation and discriminatory language patterns~\cite{gallegos2024bias}. These biases are inherited from training data and embedded in embodied interactions, potentially reinforcing stereotypes or creating differential treatment based on user demographics. 
We suggest future work to include multifaceted bias detection across embodied characteristics and natural language generation, and debiased preference learning with fair ranking that ensures equitable personalization across diverse user demographics.

\header{Safety, transparency, and accountability}
The design and development of social robots with \ac{RS} modules should integrate perspectives from both technical and socio-technical domains~\cite{sartori2022sociotechnical}.
Social robots operate in complex social environments where personalized behaviors can significantly impact human well-being and societal structures, necessitating fundamental questions around safety (potential negative effects on humans and environment), transparency (understanding robot cognition and decision-making), and accountability (responsibility when robot actions cause harm). 
We suggest future work to include interdisciplinary collaboration between technical researchers and socio-technical experts (\eg ethicists, social scientists, legal scholars, policymakers) and translating technical findings into actionable recommendations for regulatory bodies to bridge research innovations with responsible regulation.

\section{Conclusion}

In this paper, we establish the functional foundation for integrating \acfp{RS} into \acs{HRI} and social robots, aiming to leverage \ac{RS} techniques for improving personalization and user experiences.
Building on the alignment between paradigms underlying \acp{RS} and social robotics, we design a framework for social robots as \acp{RS}, which comprises three modular, plug-and-play components that ensure seamless integration, each implementing one of three key \ac{RS} techniques: user profiling, ranking, and responsible computing.
Mathematical formulations and use cases demonstrate the feasibility and efficacy of our proposed framework.
We further identify open challenges and suggest future research directions for advancing the research within this framework.
Our proposed framework has broad implications for state-of-the-art social robots, as it provides a systematic way to integrate RS intelligence into social robots, 
laying the groundwork and agenda for personalization and adaptation to diverse users in real-world deployments.

\newpage
\balance
\bibliographystyle{ACM-Reference-Format}
\bibliography{huang25_with_doi}

\end{document}